\documentclass[preprint,number,review,12pt]{elsarticle} 

\usepackage[top=2.5cm,bottom=2.5cm,left=2.5cm,right=2.5cm]{geometry}
\usepackage{amssymb}
\usepackage{hyperref}
\usepackage{graphicx}
\usepackage{fullpage}
\usepackage{mathrsfs}
\usepackage{setspace}
\usepackage{diagbox}
\usepackage{bm}
\usepackage{amssymb}
\usepackage{graphicx,subcaption}
\usepackage[normalem]{ulem}
\usepackage[dvipsnames]{xcolor}
\usepackage{comment}
\definecolor{mypink}{rgb}{0.858, 0.188, 0.478}
\usepackage{lineno}
\usepackage{verbatim}
\usepackage{xcolor}
\usepackage{eqnarray}
\usepackage[T1]{fontenc}
\usepackage{lmodern}
\usepackage{mathtools}
\RequirePackage{ifthen}
\RequirePackage{graphicx}
\usepackage{algpseudocode}
\RequirePackage{amsmath}
\RequirePackage{fancyhdr}
\usepackage{algorithm,algpseudocode}
\usepackage{booktabs}
\usepackage{fancyhdr,graphicx,amsmath,amssymb,microtype}
\usepackage{alltt}
\include{pythonlisting}
\usepackage{setspace}
\usepackage{booktabs}
\usepackage{multirow}
\usepackage{glossaries}
\newacronym{bc}{BC}{boundary condition}
\def \hb{\mathbf{h}}

\def \cb{\mathbf{c}}

\usepackage{enumitem}

\newcommand{\mcL}{\mathcal{L}}

\usepackage{xcolor}

\usepackage[utf8]{inputenc}
\usepackage[english]{babel}
\definecolor{purp}{rgb}{0.4,0.2,0.8}

\usepackage{caption}
\usepackage{subcaption}
\usepackage{hyperref}
\definecolor{custom-blue}{RGB}{3,69,173}
\hypersetup{
    colorlinks = true,
    urlcolor   = blue,
    citecolor  = custom-blue,
}
\biboptions{numbers,sort&compress}
\journal{Engineering Structures}

\begin{document}
\begin{frontmatter}
\title{Neural Operators for Stochastic Modeling of Nonlinear Structural System Response to Natural Hazards}

\author[1]{Somdatta Goswami\corref{cor1}} \ead{sgoswam4@jhu.edu}
\author[1,2]{Dimitris G. Giovanis\corref{cor1}}\ead{dgiovan1@jhu.edu}
\author[4]{Bowei Li}\ead{boweili@ttu.edu}
\author[3]{Seymour M.J. Spence\corref{cor2}} \ead{smjs@umich.edu}
\author[1,2]{Michael D. Shields} \ead{michael.shields@jhu.edu}

\address[1]{Department of Civil \& Systems Engineering, Johns Hopkins University, USA}
\address[2]{Hopkins Extreme Materials Institute, Johns Hopkins University, USA}
\address[3]{Department of Civil and Environmental Engineering, University of Michigan, USA}
\address[4]{Department of Civil, Environmental, and Construction Engineering, Texas Tech University, USA}

\cortext[cor1]{Equal Contribution.}
\cortext[cor2]{Corresponding author. The corresponding author is an editor of this journal. In accordance with policy, the corresponding author was blinded to the entire peer review process.} 

\begin{abstract}
Traditionally, neural networks have been employed to learn the mapping between finite-dimensional Euclidean spaces. However, recent research has opened up new horizons, focusing on the utilization of deep neural networks to learn operators capable of mapping infinite-dimensional function spaces. In this work, we employ two state-of-the-art neural operators, the deep operator network (DeepONet) and the Fourier neural operator (FNO) for the prediction of  the nonlinear time history response of structural systems exposed to natural hazards, such as earthquakes and wind. 
Specifically, we propose two architectures, a self-adaptive FNO and a Fast Fourier Transform-based DeepONet (DeepFNOnet), where we employ a FNO beyond the DeepONet to learn the discrepancy between the ground truth and the solution predicted by the DeepONet. To demonstrate the efficiency and applicability of the architectures, two problems are considered. In the first, we use the proposed model to predict the seismic nonlinear dynamic response of a six-story shear building subject to stochastic ground motions. In the second problem, we employ the operators to predict the wind-induced nonlinear dynamic response of a high-rise building while explicitly accounting for the stochastic nature of the wind excitation. In both cases, the trained metamodels achieve high accuracy while being orders of magnitude faster than their corresponding high-fidelity models.
\end{abstract}


\end{frontmatter}

\section{Introduction}
\label{S:Intro}

Modern probabilistic performance assessments of structural systems often require detailed simulation of the system response to prescribed hazards. Traditional simulation approaches, while accurate, demand substantial computational resources and time, often rendering them impractical for the performance estimation of large-scale structural systems subject to natural hazards characterized by stochastic models. Beyond probabilistic performance assessments of structural systems, detailed simulations become unfeasible for large-scale regional hazard modeling where a significant number of scenarios need to be considered for a large portfolio of structures. To overcome these challenges, surrogate models employing polynomial response surfaces~\cite{seo2012metamodel,perotti2013numerical,saha2016uncertainty,segura2020metamodel,goswami2016reliability}, polynomial chaos expansions~\cite{zhu2023seismic,giovanis_etal_2024}, support vector machine~\cite{gharehbaghi2019estimating,segura2020metamodel}, Kriging models~\cite{gidaris2015kriging,chatterjee2017efficient,ghosh2019kriging,shi2019adaptive}, and deep neural networks \cite{vaidyanathan2005artificial,gharehbaghi2019estimating,kim2020probabilistic,kim2019response} have been strategically designed to approximate the responses of complex models, promising significant reductions in computational cost without substantially sacrificing accuracy. The recently introduced metamodeling techniques based on the autoregressive frameworks \cite{spiridonakos2015metamodeling,mai2016surrogate,Bhattacharyya20,li2021response} and deep learning sequence-to-sequence mapping approaches \cite{kundu2020deep,zhang2020physics,zhang2020physicscnn,wang2020knowledge,li2022metamodeling,Li_2024,Atila_2025} are capable of predicting the entire time histories response of nonlinear dynamic structural systems. The ability of metamodels to provide quick and reliable estimates of system response time histories to support comprehensive performance analyses is invaluable for propagating uncertainty and estimating system-level probabilistic performance metrics. The advent of machine learning (ML) technologies marks a potentially transformative shift in metamodeling, enhancing the predictive analysis of system responses to natural hazards.

In the emerging field of metamodeling, neural operators represent a particularly promising approach to capturing the intricate effects of natural hazards on structural systems. Unlike traditional metamodeling methods that define a mapping between finite dimensional vector spaces, neural operators learn mappings between infinite-dimensional function spaces, which have the potential to better represent the time-varying structural response to complex hazards. This potentially enables more accurate predictions of system behavior under a more diverse set of conditions. The adaptability and scalability of neural operators make them particularly suited for modeling the stochastic nature of the response of structural systems to natural hazards, potentially providing a robust tool for engineers to rapidly predict the vulnerabilities of complex structural systems and optimize their design for resilience. These techniques not only hold the promise of accelerating the assessment process of nonlinear stochastic systems but also improving the feasibility of conducting extensive scenario analyses. In this work, we employ two neural operators, the Deep Operator Network (DeepONet)~\cite{lu2021learning} and the Fourier Neural Operator (FNO)~\cite{li2020fourier}, and explore additional architectures combining these two operators to learn the response of nonlinear structural systems subject to seismic and wind hazards characterized by general stochastic models. 

The paper is organized as follows. In Section~\ref{S:NO}, we describe operator learning and the key ingredients of the DeepONet and FNO architectures. In Section~\ref{S:NO}, we present two extensions of the DeepONet and FNO architectures for stochastic modeling of system response to natural hazards. The first is based on the self-adaptation of the DeepONet/FNO weights during training. The second scheme leverages an integrated DeepONet-FNO framework (termed DeepFNOnet). In Section~\ref{S:examples}, we showcase the performance of the proposed methods for earthquake and wind engineering problems. In the first example, we employ the operators to predict the seismic nonlinear dynamic response of a six-story shear building subjected to stochastic ground acceleration. In the second problem, we predict the response of a 37-story building in the presence of extreme winds. Finally, we summarize our observations and provide concluding remarks.


\section{Neural Operator Learning Frameworks}
\label{S:NO}

Operator learning is a powerful framework for constructing metamodels for physical systems governed by differential equations. The framework recognizes that physical laws are fundamentally described through mathematical operators, and aims to use data generated from physical processes to learn these operators using ML methods. Neural operators are a family of operator learning methods that use neural networks to learn a function that approximates the operator. There are numerous neural operators, which can be categorized into meta-architectures. One category is those derived from the universal approximation theorem for operators~\cite{chen1995universal} that include the DeepONet~\cite{lu2021learning} and embedding operators such as the Resolution Independent Neural Operator (RINO)~\cite{bahmani2024resolution} and Basis-to-Basis (B2B) neural operator~\cite{ingebrand2025basis}, among others. A second category are the integral operators that include the Fourier neural operator (FNO)~\cite{li2020fourier}, wavelet neural operator (WNO)~\cite{tripura2023wavelet}, graph kernel network (GKN)~\cite{anandkumar2020neural}, convolutional neural operator~\cite{raonic2023convolutional}, and Laplace neural operator (LNO)~\cite{cao2024laplace}. In this work, we focus on two specific neural operators: DeepONet and FNO. 

In this section, we first establish the mathematical framework for the operator learning task for nonlinear dynamic structural response to natural hazards investigated in this study. Subsequently, we describe the architectural details of the basic (vanilla) DeepONet and FNO models. We introduce two novel architectures: (1) self-adaptive FNO (SA-FNO), which extends the concept of self-adaptive DeepONet \cite{kontolati2022influence} by adaptively penalizing individual time points in response signals; and (2) DeepFNOnet, which enhances DeepONet's performance for this class of problems by sequentially coupling it with FNO.


\subsection{Problem Statement}

The nonlinear response of a multi-degree-of-freedom structural system subjected to stochastic excitation from a natural hazard is governed by the general equation of motion given by:
\begin{equation}
    \mathbf{M} \ddot{\mathbf{u}}(t) + \mathbf{f}_{\text{D}}(\mathbf{u}(t), \dot{\mathbf{u}}(t))(t) + \mathbf{f}_{\text{NL}}(\mathbf{u}(t), \dot{\mathbf{u}}(t))(t) = \mathbf{F}(t)
    \label{eqn:EOM}
\end{equation}
where $\mathbf{u}(t) = \{u_1(t), u_2(t), \ldots, u_N(t)\}^\intercal$ is the vector of continuous time displacements at each of the $N$ degrees of freedom of the discretized structural system; $\mathbf{M}$ is the $N\times N$ mass matrix of the system; $\mathbf{f}_{\text{D}}(t)$ is the $N\times 1$ damping force which in general will depend on the displacement, $\mathbf{u}(t)$, and velocity, $\dot{\mathbf{u}}(t)$, response of the system; $\mathbf{f}_{\text{NL}}(t)$ is the vector of nonlinear restoring forces that in general will depend on $\mathbf{u}(t)$ and $\dot{\mathbf{u}}(t)$; and $\mathbf{F}(t) = \{f_1(t), f_2(t), \ldots, f_N(t)\}^\intercal$ is the vector of continuous-time stochastic excitation at each of the $N$ degrees of freedom from the specified natural hazard. 

In the context of operator learning, let $\mathcal{G}(\mathbf{F}(t))(t)$ denote the non-linear operator in Eq.~\eqref{eqn:EOM} that maps the stochastic input function, $\mathbf{F}(t)$, to the corresponding output response function, $\mathbf{u}(t)$, such that the input and the output spaces are defined on a bounded time domain $[0, T]$. Examples of input functions include stochastic wind or seismic excitation, while the corresponding output functions are, as previously stated, displacements (including absolute values or inter-story drifts). Neural operators employ deep neural networks to approximate $\mathcal{G}$ by the functional $\mathcal{G}_{\bm\theta}$, which is parametrized by a set of learnable parameters $\bm{\theta} \in \bm{\Theta}$ such that $\mathcal{G}(\mathbf{F}(t))(t)\approx \mathcal{G}_{\bm\theta^*}(\mathbf{F}(t))(t)$ where $\bm{\theta}^*$ are the optimal set of parameters.

\subsection{Deep Operator Network (DeepONet)}
\label{S:DeepOnet}

\begin{figure}[]
\begin{center}
\includegraphics[width=1\textwidth]{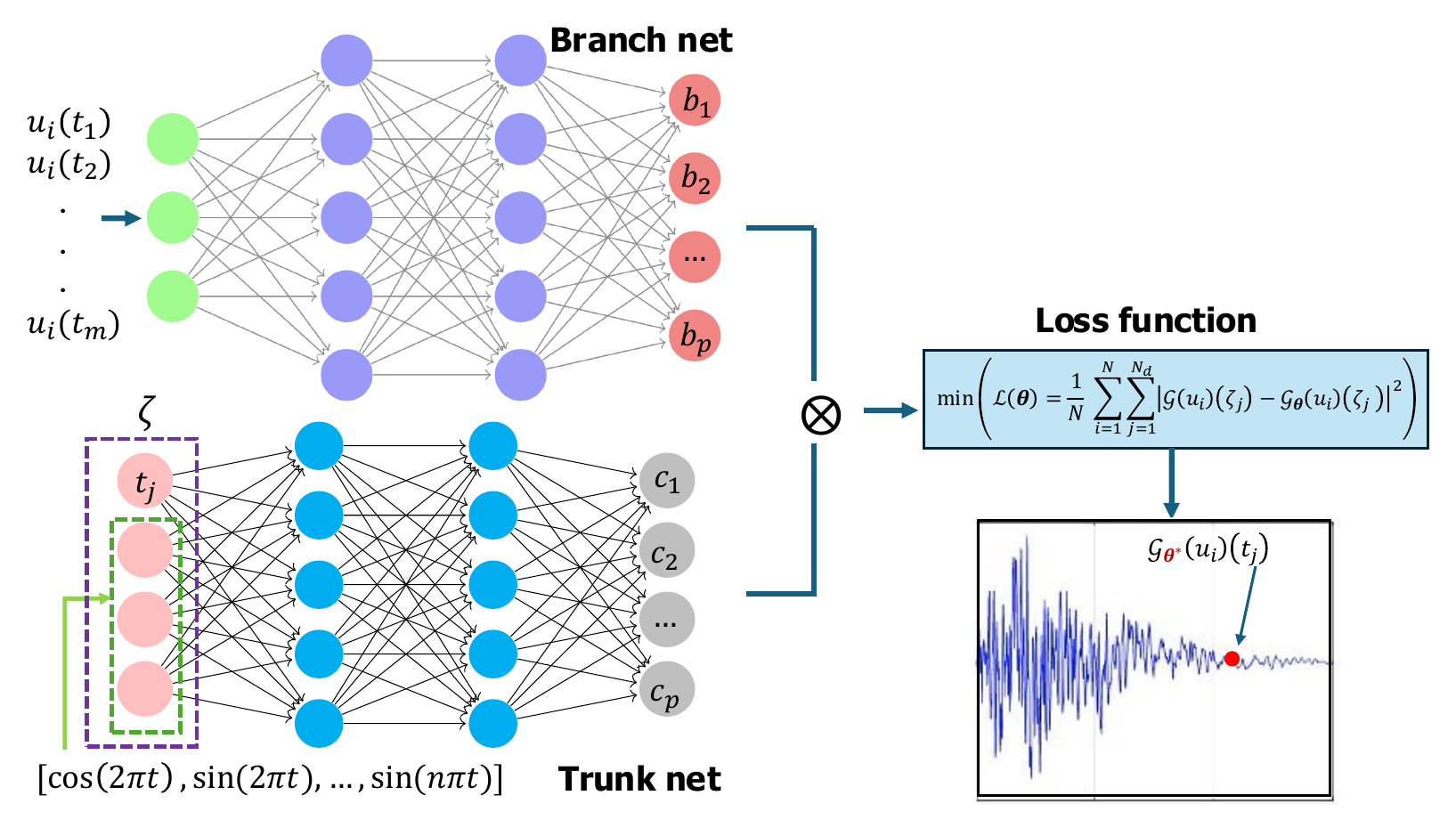}
\caption{DeepONet with an FNN architecture for both the branch and the trunk networks. The element-wise dot product of the output feature embedding of the branch net, $[b_1, b_2, \ldots, b_p]^\intercal \in \mathbb{R}^p$, and the trunk net, $[c_1, c_2, \ldots, c_p]^\intercal \in \mathbb{R}^p$, yields the solution operator, $\mathcal G_{\bm\theta}$, where $\bm{\theta}$ denotes the trainable parameters of the network. The loss function, $\mathcal L(\bm{\theta})$, which is obtained as the difference between the exact and predicted solution over all spatial and temporal points, is minimized to obtain the optimized parameters of the network, $\bm{\theta}^{*}$. For any point $\zeta=\{x_j, y_j, z_j, t_j\}$, the output $\mathcal{G}(u)(\zeta) \in \mathbb{R}$ is a real number.}
\label{fig:DeepONet}
\end{center}
\end{figure}

The DeepONet, illustrated in Figure~\ref{fig:DeepONet}, consists of two deep neural networks: a branch network and a trunk network. The branch network processes the input functions $\mathbf{F}_i(t_j)$, where $i= 1, 2, \ldots, M$ and $M$ is the total number of training samples evaluated at $n_t$ time steps, $t_j$ with $j =1, 2, \ldots, n_t$. The branch network is not restricted to any particular architecture, hence could take a convolutional neural network (CNN), or fully-connected feed-forward neural network (FNN), or a recurrent neural network (RNN). The trunk network defines a continuous basis for the solution, which in general takes as input the continuous spatio-temporal coordinates $\boldsymbol{\zeta} =\{x, y, z, t\} \in \mathbb{R}^{N_d}$, where $N_d=d_x \times d_y \times d_z \times n_t$, with $d_x$, $d_y$, and $d_z$ defining the spatial discretization and $n_t$ denoting the temporal discretization. In the applications considered here, the structural degrees of freedom are considered located on each floor. We therefore do not consider interpolation of displacements between floors. Consequently, the trunk network in this application consists only of temporal inputs, $\boldsymbol{\zeta}=t$, and predictions are only made at the prescribed floor levels. Usually, the trunk net adopts an FNN architecture.  

The outputs of the branch and trunk networks, i.e., $[b_1, b_2, \ldots, b_p]^\intercal \in \mathbb{R}^p$ and $[c_1, c_2, \ldots, c_p]^\intercal \in \mathbb{R}^p$ where $p \in \mathbb{Z}^+$ is the size of the feature space, are then combined through an inner product to produce the final output: $\mathcal G_{\bm\theta}(\mathbf{F})(t) \in \mathbb{R}$ where $\bm{\theta} = \left(\mathbf{W}, \mathbf{b}\right)$ consists of the trainable weights, $\mathbf{W}$, and biases, $\mathbf{b}$, of the two networks. The solution operator, $\mathcal G_{\bm\theta}(\mathbf{F})(t)$, is defined as:
\begin{equation} \label{eq:solution_deeponet}
\mathcal{G}_{\bm{\theta}}(\mathbf{F})(t)= \sum_{i=1}^p b_i \cdot c_i = \sum_{i=1}^p b_i(\mathbf{F}(t_1), \ldots, \mathbf{F}(t_{n_t})) \cdot c_i(t)
\end{equation}
The parameters $\bm{\theta}$ in Eq.~\eqref{eq:solution_deeponet}  are obtained by minimizing the loss function:
\begin{equation}
    \mathcal L(\bm{\theta}) = \frac{1}{M} \sum_{i=1}^{M} \sum_{j=1}^{n_t} | \mathcal{G}(\mathbf{F}_i)(t_j)- \mathcal{G}_{\boldsymbol{\theta}}(\mathbf{F}_i)(t_j)|^2,
\end{equation}
where $M$ is the number of samples while $\mathcal{G}(\mathbf{F}_i)(t_j)$ and $\mathcal{G}_{\boldsymbol{\theta}}(\mathbf{F}_i)(t_j)$ represent the exact and the predicted solutions, respectively. 

An extension of DeepONet, namely, DeepONet with self-adaptivity (SA-DeepONet), was introduced in \cite{kontolati2022influence}. This approach, motivated by the framework proposed in \cite{mcclenny2020self}, replaces the constant and manually modulating penalty parameters, $\bm\lambda$, with trainable parameters. The basic idea behind self-adaptivity is to make the penalty parameters increase where the corresponding loss is higher, which is accomplished by training the network to simultaneously minimize the losses and maximize the value of the penalty parameters. The modified loss function employed here is defined as:
\begin{equation}
    \mathcal L(\bm \theta, \bm \lambda) = \frac{1}{M}\sum_{i = 1}^{M} \sum_{j=1}^{n_t}g(\lambda_j)|\mathcal{G}(\mathbf{F}_i)(t_j)- \mathcal G_{\bm\theta}(\mathbf{F}_i)(t_j)|^2,
\end{equation}
where $\bm \lambda = \{\lambda_1, \lambda_2, \cdots \lambda_j\}$ are $n_t$ self-adaptive parameters, each associated with an evaluation point, $t_j$. These parameters are constrained to increase monotonically and are always positive. The modified objective function is then defined as:
\begin{equation}
    \min_{\bm \theta} \max_{\bm \lambda}\mathcal L(\bm \theta, \bm \lambda).
\end{equation}
The self-adaptive weights are updated using the gradient ascent method, such that:
\begin{equation}
    \lambda_j^{k+1} = \lambda_j^{k} + \eta_{\lambda}\nabla_{\lambda_j}\mathcal L(\bm \theta, \bm \lambda),
\end{equation}
where $\eta_{\bm \lambda}$ is the learning rate of the self-adaptive weights with:
\begin{equation}
    \nabla_{\lambda_j}\mathcal L = \left[g'(\lambda_j)(\mathcal{G}(\mathbf{F}_i)(t_j)- \mathcal G_{\bm\theta}(\mathbf{F}_i)(t_j))^2\right]^\intercal.
\end{equation}
Therefore, if $g'(\lambda_j)>0$, $\nabla_{\lambda_j}\mathcal L$ would be zero only if the term $(\mathcal{G}(\mathbf{F}_i)(t)- \mathcal G_{\bm\theta}(\mathbf{F}_i)(t))$ is zero. Implementing self-adaptive weights in \cite{kontolati2022influence} has considerably improved the accuracy prediction of discontinuities or non-smooth features in the solution.


\subsection{Fourier Neural Operator (FNO)}
\label{S:FNO}

The FNO~\cite{li2020fourier} is an integral neural operator that replaces the kernel integral operator with a convolution operator defined in the Fourier space. The operator takes input functions defined on a well-defined, equally spaced lattice grid and outputs the field of interest on the same grid points. The network parameters are defined and learned in the Fourier space rather than in the physical space, i.e., the coefficients of the Fourier series of the output function are learned from the data.

\begin{figure}[]
    \centering
    \includegraphics[width=\textwidth]{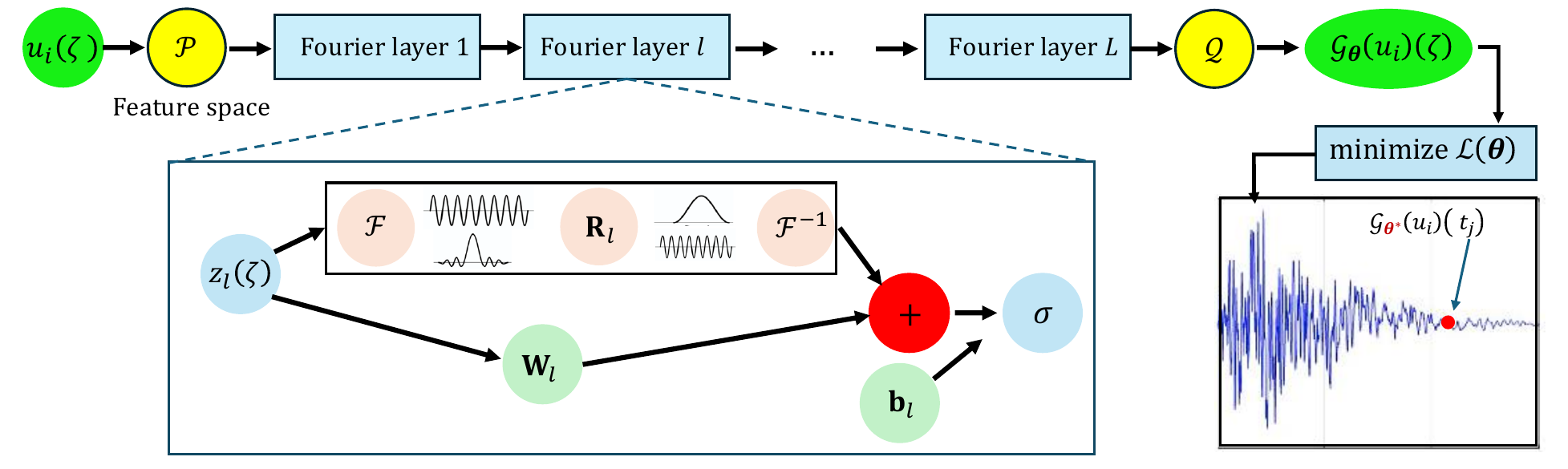}
    \caption{Schematic of the FNO~\cite{li2020fourier}. The FNO takes uniformly discretized temporal inputs and associated features, such as ground motions in seismic engineering problems or dynamic wind loads in wind engineering problems, at each floor level of the discretized building system. These inputs are first projected into a higher-dimensional feature space, $\mathcal{P}$, via a shallow neural network. The transformed representation is then processed through multiple Fourier layers, each comprising a forward Fast Fourier Transform (FFT), a linear transformation of low-frequency modes, and an inverse FFT. After each layer, the output is augmented with a learned weight matrix and passed through an activation function to introduce nonlinearity.  Vectorized outputs are enabled through the use of multiple neurons in the shallow neural network, $\mathcal Q$.}
    \label{fig:fno}
\end{figure}

A schematic representation of the FNO is shown in Figure \ref{fig:fno}. The FNO has three components. First, the input function, $\mathbf{F}_i(t)$, is lifted to a higher-dimensional representation, $\hb(t,0)$, in the feature space, $\mathcal{P}$, through a lifting layer, which is often defined by a parameterized linear transformation or a shallow neural network. Then, a sequence of $L$ Fourier layers sequentially performs the following nonlinear operation on the lifted representation:
\begin{align}\label{eq:FNO}
\hb(t,j+1)=&\mcL_j^{FNO}[\hb(t,j)]\\
:=&\sigma\left(W_j\hb(t,j)+\mathcal{F}^{-1}[R_j\cdot \mathcal{F}[\hb(t,j)]](t)+ \cb_j\right), \quad j = 0,\dots,L-1.
\end{align}
where $\sigma$ denotes the activation function while $W_j$, $\cb_j$, and $R_j$ are trainable parameters for the $j$-th layer (each layer has different parameters, i.e., different kernels, weights, and biases). Lastly, the output, $\mathcal G_{\bm \theta}(\mathbf{F}_i)(t)$, is obtained by projecting $\hb(x,L)$ through a local transformation operator layer, $Q$. While traditional FNO implementations typically produce scalar outputs, we have employed vectorized outputs by using multiple neurons in the shallow neural network, $\mathcal Q$, to simultaneously capture responses across multiple floor levels in a single training iteration.

Motivated by the success of SA-DeepONet, we employ a similar self-adaptivity scheme for FNO, termed SA-FNO. In this work, with the self-adaptivity architecture, we associate every temporal point on the output space with a trainable weight parameter, which adapts during the learning process to the temporal dynamics of the response. 

\subsection{DeepONet integrated with FNO: DeepFNOnet}
\label{S:DeepFNOnet}

Because the FNO establishes the operator through a sequence of Fourier transforms of the input -- effectively building the operator partially in the frequency domain -- the FNO demonstrates robust and accurate performance in capturing long-time horizon responses for systems with high-frequency content. DeepONet, on the other hand, exhibits significant limitations in generating accurate predictions for extended high-frequency temporal dynamics because of spectral bias. To improve its performance for long-time predictions, we introduce an extension to the DeepONet, termed the DeepFNOnet, aimed at overcoming this spectral bias. 

\begin{figure}[t!]
    \centering
    \includegraphics[width=\textwidth]{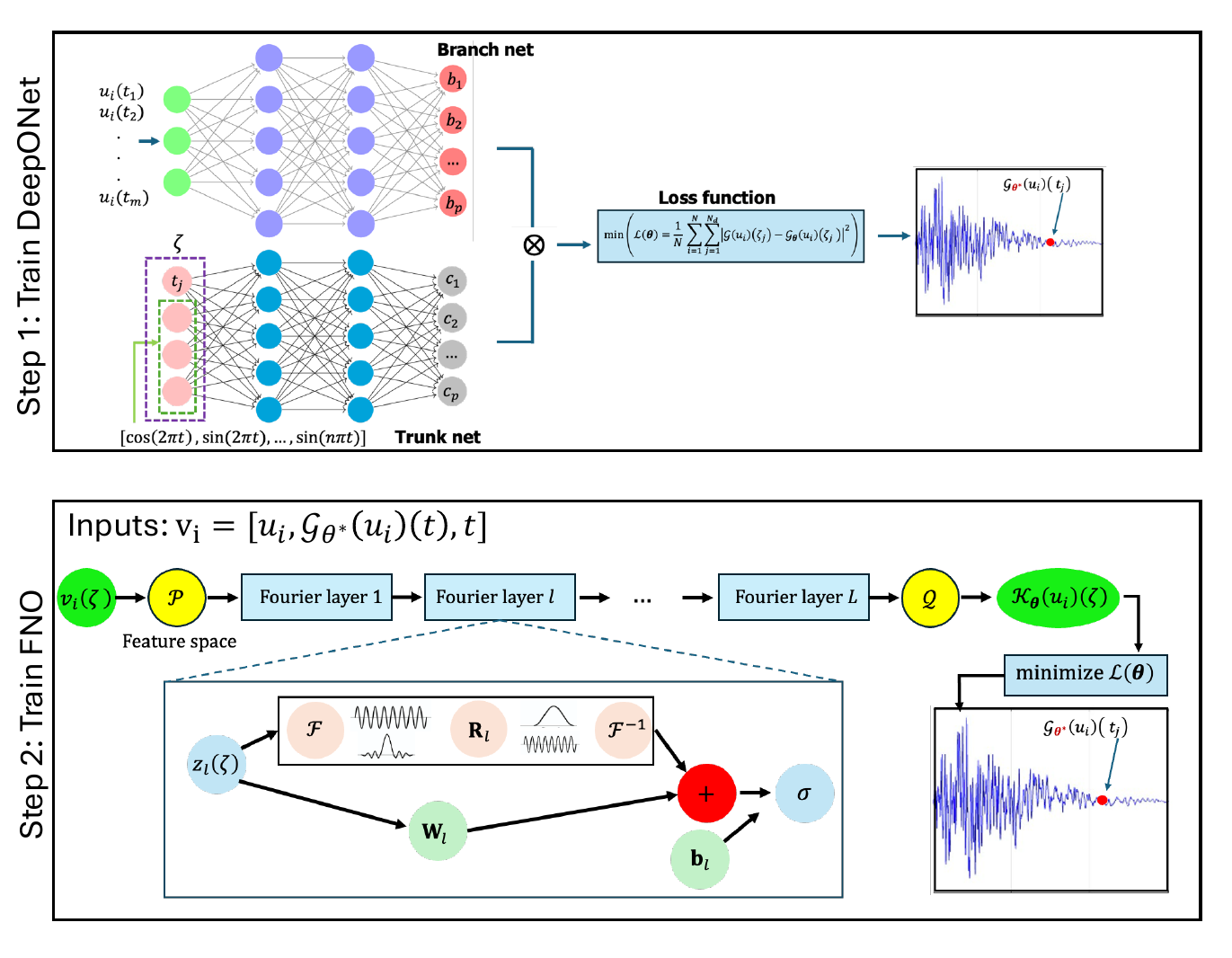}
    \caption{Schematic representation of DeepFNOnet, a hybrid neural architecture employing a two-stage training methodology. The framework first follows the conventional DeepONet training procedure, subsequently leveraging DeepONet's predictions by concatenating them with the original input time signals as enhanced inputs to the FNO. This integrated approach enables dual capabilities: temporal interpolation through the DeepONet framework and one-shot super-resolution facilitated by the FNO architecture, thereby synergistically combining the strengths of both neural operators.}
    \label{fig:DeepFNOnet}
\end{figure}

The DeepFNOnet framework mitigates spectral bias through a two-stage training approach, illustrated in Figure~\ref{fig:DeepFNOnet}, that synergistically combines the strengths of DeepONet and FNO. By first training a DeepONet to capture fundamental operator transformations and then using its predictions as enhanced inputs to an FNO, the framework ensures higher predictive accuracy than DeepONet alone. While one could argue that the predictive accuracy of the integrated framework might be enhanced through concurrent training of the networks, our current implementation maintains a sequential training approach. This decision stems from computational constraints, as all framework components were trained on a single GPU, and simultaneously updating a large number of trainable parameters across multiple networks would pose significant computational challenges. We aim to address this limitation in the future with distributed training and by developing Fourier feature-enhanced architectures to learn the high-frequency components.

%


\section{Numerical Examples}
\label{S:examples}

\subsection{Preamble}
\label{S:pream}

In this section, we analyze two distinct problems: the seismic response of a six-story building subjected to stochastic earthquake loading, and the dynamic response of a 37-story building subjected to stochastic wind excitation. We apply the previously outlined neural operator architectures to both scenarios and compare their performance. We report the accuracy using mean squared error (MSE), relative $\mathcal{L}_2$, and absolute point-wise errors ($e$), defined, respectively, as follows: 
%
\begin{equation}
\left\{ 
\begin{array}{lll}
    \text{MSE} = \frac{1}{N_{test}}\sum_{i=1}^{N_{test}}(y_i - \hat{y}_i)^2\\
    \text{Relative}\;\mathcal{L}_2 = \frac{\sqrt{\sum_{i=1}^{N_{test}} (y_i - \hat{y}_i)^2}}{\sqrt{\sum_{i=1}^{N_{test}} (y_i)^2}}\\
    e = |y_i - \hat{y}_i|
\end{array}\right.  
\label{eqn:err_mesur}  
\end{equation}
where $y_i$ denotes the ground truth generated using the high-fidelity model, $\hat{y}_i$ is the prediction obtained from the neural operator, and $N_{test}$ is the total number of testing samples.


\subsection{Seismic Response of a Shear Building}
\label{S:eq}


\subsubsection{Model Setup}
\label{S:buildsetup}

\begin{figure}[tb]
    \centering
    \includegraphics[width=0.75\textwidth]{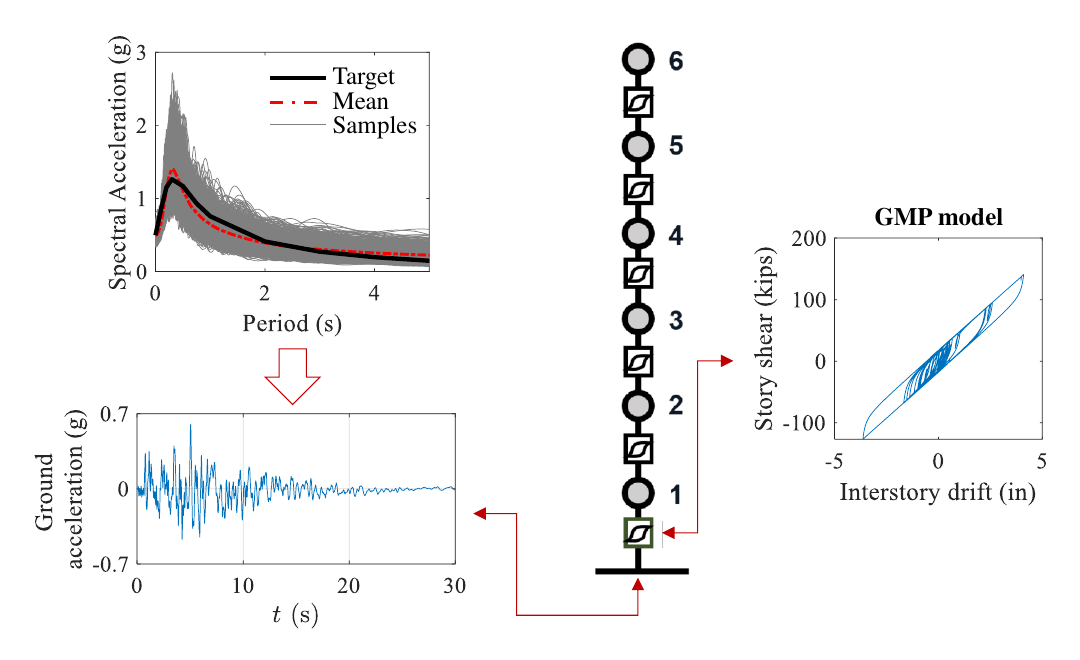}
    \caption{Schematic representation of the 6-story nonlinear shear building subjected to stochastic ground motions.}
    \label{fig:6story}
\end{figure}

The first case study examines the seismic response of a six-story nonlinear shear building (see Fig.~\ref{fig:6story}) designed to represent a typical mid-rise steel structure. The response of the structure to the imposed seismic base excitation is governed by the following equations of motion:
\begin{equation}
\textbf{M}\ddot{\textbf{u}}(t) +\textbf{C}\dot{\textbf{u}}(t) + \textbf{f}_{\text{NL}}(\textbf{u}(t),\dot{\textbf{u}}(t))(t)=-\textbf{M}\{\mathbf{1}\}\ddot{u}_g(t),
\label{eqn:Seismic_EOM}
\end{equation}
where $\textbf{M}$ is the $6\times 6$ diagonal mass matrix, with diagonal terms assembled from mass calculated from the floor weights of Table \ref{tab:params}; $\textbf{C}$ is a $6\times 6$ Rayleigh damping matrix obtained through the linear combination of the mass matrix and initial stiffness matrix while imposing a modal damping ratio of 0.05 at the 1st and the 2nd natural frequencies; $\textbf{f}_{\text{NL}}(\textbf{u}(t),\dot{\textbf{u}}(t))(t)$ is the non-linear restoring force of the structure; $\{\mathbf{1}\}$ is a $6\times 1$ vector of ones; and $\ddot{u}_g(t)$ is the horizontal ground acceleration modeled as a filtered white noise stochastic process (described later). The target response is determined by solving Eq.~\eqref{eqn:Seismic_EOM} utilizing the average constant acceleration integration scheme in conjunction with the Newton-Raphson method. A time step of $\Delta t = 0.005$ s was implemented, with the response estimated over a total duration of $T = 30$ s, resulting in 6001 time steps.

The building was modeled as a shear structure, which by definition treats the floors as rigid diaphragms while neglecting the axial deformability of the columns. Furthermore, consistent with this assumption, the floors and the ground provide fully fixed support to the columns, preventing any rotation at the floor level. The shear structure was modeled in OpenSees \cite{opensees} with the interstory load displacement relationship modeled using a Giuffre-Menegotto-Pinto (GMP) model~\cite{giuffre1970comportamento,menegotto1973method,carreno2020material}. The parameters defining the building model, including the GMP model, are reported in Table \ref{tab:params}.  

\begin{table}[]
\centering
\footnotesize
\caption{Model parameters of the 6-story shear building.}
\begin{tabular}{l|l|l}
\hline
\textbf{Parameter} & \textbf{Description} & \textbf{Value} \\
\hline
\( h \) & Story height & 100 in\\
\( w \) & Floor weight & 60 kips \\
\( K_e \) & Initial Stiffness & 300 kip/in \\
\( \alpha \) & Post-yield stiffness ratio & 0.1 \\
\( f_y \) & Yield strength & 20 kip/in$^2$ \\
\( R_0\) & Elastic--Plastic Transition Parameter & 15 \\
\hline
\end{tabular}

\label{tab:params}
\end{table}

\subsubsection{Stochastic Ground Motion Model}
\label{S:BuildMod}

The ground motion model uses filtered white noise to simulate seismic excitation. This involves passing a white noise signal (a random signal with a consistent spectral density) through a filter specifically designed to replicate the frequency characteristics of real ground motion \cite{rezaeian2010simulation}. By filtering the white noise, the resulting signal maintains its randomness while acquiring features akin to those of genuine seismic events, including dominant frequencies and variations in amplitude.

The model is defined by the following six parameters: Arias intensity, $\bar{I}_a$; effective duration, $D_{5-95}$ (defined as the time interval between the time points when 5\% and 95\% of $\bar{I}_a$ are reached);  time point when 45\% $\bar{I}_a$ is reached, $t_\text{mid}$; filter circular frequency when 45\% $\bar{I}_a$ is reached, $\omega_\text{mid}$; rate of change of the circular filter frequency, $\omega'$; and filter damping ratio $\zeta_\text{f}$. Together these parameters govern the time and frequency domain characteristics of the stochastic ground motion model. The seismic hazard for this example was defined as the target spectrum associated with a 10\% chance of exceedance in 50 years for the Loma Prieta region. This was generated using tools provided by the United States Geological Survey (USGS) and is illustrated in the top left of Fig.~\ref{fig:6story}. The parameters of the stochastic ground motion model were calibrated to reproduce, in the mean sense, the target spectrum. In particular, the parameters $\bar{I}_a$, $D_{5-95}$, $t_\text{mid}$, and $\omega'$ were estimated from aground acceleration records corresponding to the 1989 Loma Prieta earthquake with a Moment magnitude of 6.93 and rupture distance of 18.3 km. The remaining parameters were calibrated to minimize the mean square error between the mean response spectrum of the stochastic  ground motion model and the target spectrum. This calibration resulted in: $\bar{I}_a=0.045$; $D_{5-95}=12.62$ s; $t_\text{mid}=4.73$ s; $\omega_\text{mid}=2\pi\times3.27$ rad/s; $\omega'=-2\pi\times0.08$ rad/s; and $\zeta_\text{f}=0.48$.

\subsubsection{Results}
\label{S:ResulS}

The ground motion model of Section \ref{S:BuildMod} was used to create a dataset of 1000 ground acceleration records. Corresponding displacement time histories at each of the six floors were generated using the model setup described in Section \ref{S:buildsetup}. Of these realizations, 800 were used for training and 200 for validation, with model performance assessed using the MSE and relative $\mathcal L_2$ error metrics defined in Eq. (\ref{eqn:err_mesur}). As described above, the ground motions were simulated for 30 s with a time step of $5\times10^{-3}$ s therefore leading to a total of 6001 discrete time points. The input to the neural operator is a $5980 \times 1$ vector which considers the entire ground motion excluding the first 21 time points which are close to zero. For the first experiment, we analyze the accuracy of the neural operators, specifically FNO and SA-FNO, to predict the response across all six floors of the building. To that end, the output from the neural operator is a matrix of size $5980 \times 6$, where each matrix column represents a floor.

\begin{figure}[]
\begin{center}
\includegraphics[trim={1cm 0cm 0cm 0cm},clip, width=1.0\textwidth]{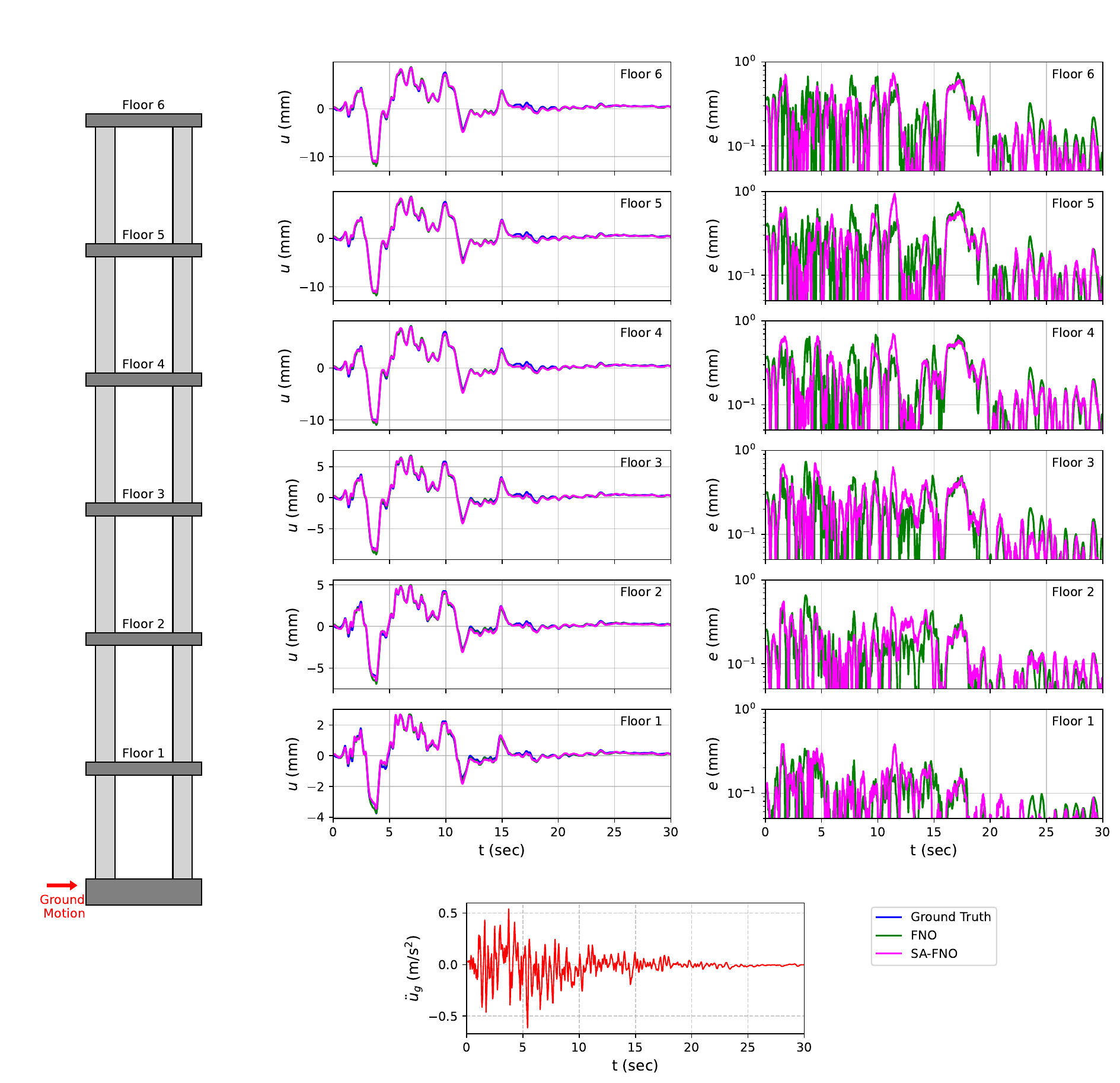}
\caption{Comparative illustration of the prediction accuracies of FNO and SA-FNO for each floor of the shear building example. The left-hand panel illustrates the building setup. The middle panel reports the comparison of the floor responses predicted by the FNO and SA-FNO to the ground truth. The right panel displays the error measure, $e$, plotted in logarithmic scale on the vertical axis. The bottom plot shows the applied base acceleration time history.}
\label{fig:EQ_sample0}
\end{center}
\end{figure}

Figure~\ref{fig:EQ_sample0} illustrates the neural operator performance for a representative sample modeled using the FNO and SA-FNO. Both frameworks successfully map ground acceleration to structural responses, with the plots in the center showing a strong match between predicted and simulated responses on each floor. The plots on the right depict the absolute error, $e$, calculated from the predicted response and the response simulated using the Newton-Raphson solver of Section \ref{S:buildsetup}, on a logarithmic scale. These plots illustrate that the SA-FNO achieves notably smaller errors on the lower floors, with the magenta line indicating errors as low as $10^{-5}$. In contrast, for the higher floors, the errors for both FNO and SA-FNO are comparable, as evidenced by the overlapping green and magenta lines.

Since the top floor typically experiences the maximum displacement during an earthquake, we focus our subsequent analysis on the prediction of the seismic response of this floor. Performance metrics for all models are reported in Table~\ref{tab:performance_metrics_models_eq}. The results show that the standalone DeepONet struggles with long-time-duration predictions, producing high errors. However, coupling DeepONet with FNO significantly improves performance, greatly reducing prediction errors. Meanwhile, the overall performance metrics of FNO and SA-FNO show similarly high accuracy. 

\begin{figure}[h!]
\begin{center}
\includegraphics[width=1.0\textwidth]{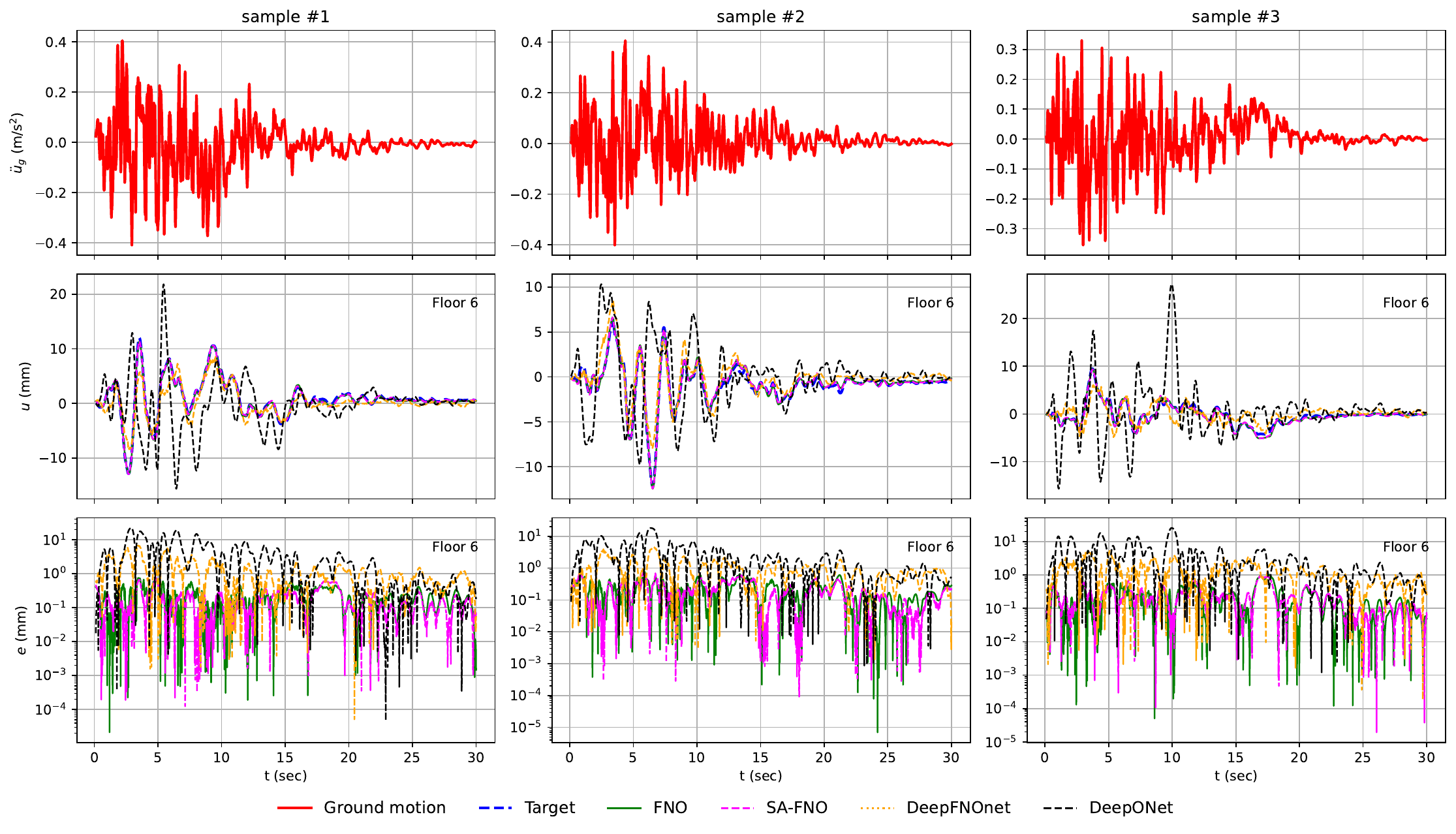}
\caption{A comparison of accuracy among all neural operators (FNO, SA-FNO, DeepFNOnet, and DeepONet) in predicting the response of the top floor. The plot illustrates the displacement response and absolute error metrics on the log scale for three representative realizations of the ground motion model.}
\label{fig:EQ_samples_topmostfloor}
\end{center}
\end{figure}

Figure~\ref{fig:EQ_samples_topmostfloor} reports the responses of three representative samples using each of the tested neural operators, highlighting the accuracy of FNO and SA-FNO for the simulated dataset. The error plots in the bottom row demonstrate that both FNO and SA-FNO outperform other neural operators, with SA-FNO exhibiting lower errors at multiple time points. However, determining the superior architectures between FNO and SA-FNO requires careful interpretation of Table~\ref{tab:performance_metrics_models_eq}. While FNO achieves the lowest MSE, SA-FNO has the lowest relative $\mathcal L_2$ error. These aggregate performance metrics, though valuable for overall comparison, may not fully capture the nuanced temporal dynamics of the response.

\begin{table}[]
\centering
\footnotesize
\caption{Performance evaluation for all considered NOs in predicting the seismic response of the top floor. Predictive accuracy of the NO architectures in terms of the MSE, best and worst sample performances, and relative $\mathcal{L}_2$ error metrics. The lowest error in each section is marked with \textbf{bold} text.}
\renewcommand{\arraystretch}{1.0} 
\setlength{\tabcolsep}{6pt} 
\begin{tabular}{l|ccc|c}
\hline
\multirow{3}{*}{\textbf{Model}} & \multicolumn{3}{c|}{\textbf{MSE}} & \textbf{Relative $\boldsymbol{\mathcal{L}_2}$} \\
               & \textbf{Overall} & \textbf{Best case} & \textbf{Worst case} & \textbf{Error} \\
                & ($\boldsymbol{\times 10^{-1}}$) & ($\boldsymbol{\times 10^{-2}}$) & ($\boldsymbol{\times 10^{1}}$) & ($\boldsymbol{\times 10^{-1}}$) \\
\hline
\textbf{FNO}            & \textbf{0.7146} & \textbf{0.2876} & \textbf{2.042} & 2.605 \\
\textbf{SA-FNO}         & 0.7816 & 0.2834 & 2.203 & \textbf{2.542} \\
\textbf{DeepFNOnet}   & 18.46 & 4.783 & 72.61 & 8.064 \\
\textbf{DeepONet}       & 329.74 & 88.03 & 1112.04 & 19.04 \\
\hline
\end{tabular}
\label{tab:performance_metrics_models_eq}
\end{table}


\subsection{Dynamic Wind Response of a Steel Frame}
\label{S:wind}

\subsubsection{Model Setup}
\label{W:buildsetup}

Modern wind assessments generally employ advanced probabilistic schemes to estimate metrics related to structural response by propagating uncertainties associated with stochastic wind excitation through sophisticated finite element models \cite{ouyang2021performance,chuang2022framework,steelcollapse2022,Li2023}. Considering Rayleigh damping, for this application, Eq. (\ref{eqn:EOM}) can be cast as:   
\begin{equation}
\textbf{M}\ddot{\textbf{q}}(t) +\textbf{C}\dot{\textbf{q}}(t) + \textbf{f}_{\text{NL}}(\textbf{q}(t),\dot{\textbf{q}}(t))(t)=\textbf{F}(v_H, \alpha)(t)
\end{equation}
where the $\textbf{q}(t)$, $\dot{\textbf{q}}(t)$, and $\ddot{\textbf{q}}(t)$ represent the displacement, velocity and acceleration response of the system; $\textbf{C}$ is the Rayleigh damping matrix; $  \textbf{f}_{\text{NL}}(\textbf{q}(t),\dot{\textbf{q}}(t))(t)$ is the nonlinear restoring force; and $\textbf{F}(v_H, \alpha)(t)$ is the external stochastic wind excitation vector calibrated to a maximum mean hourly wind speed, $v_H$, and specified direction, $\alpha$.

\begin{figure}[]
    \centering
    \includegraphics[width=0.45\textwidth]{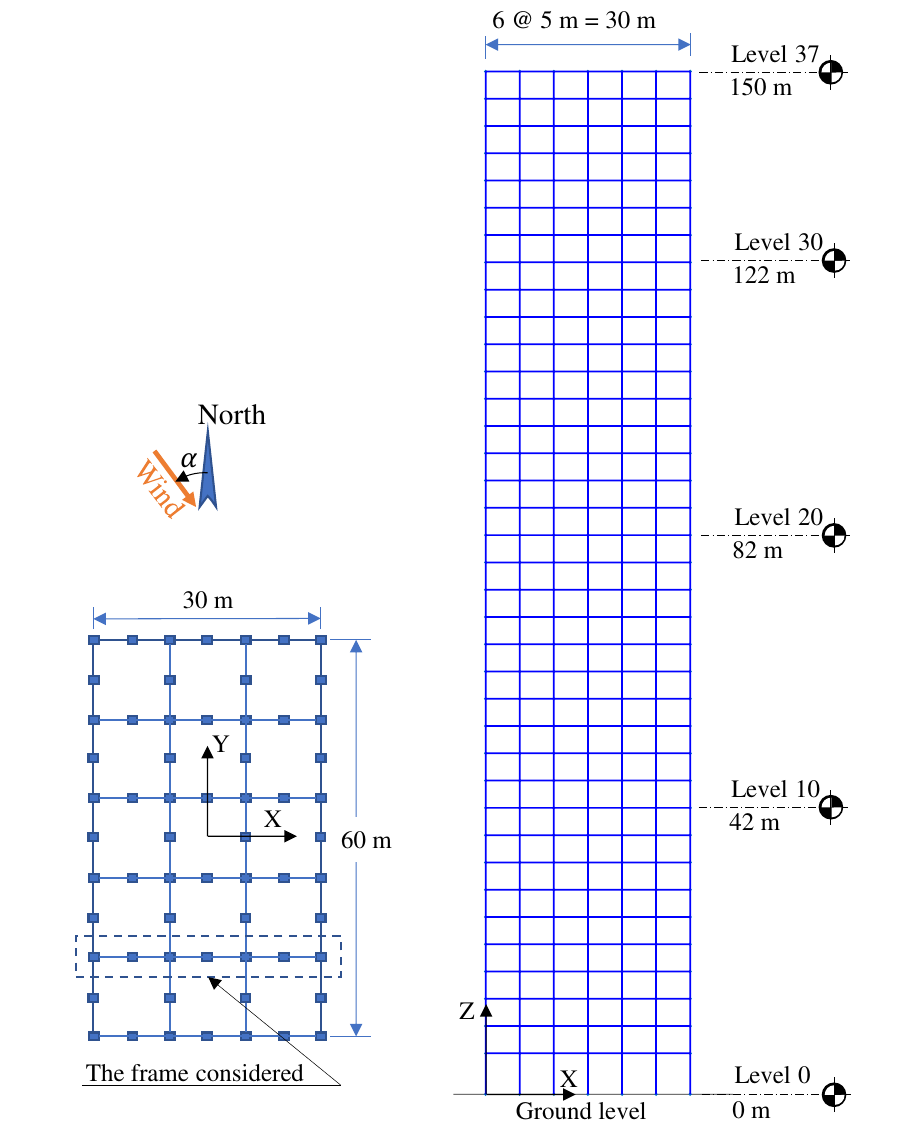}
    \caption{Schematic Representation of the 37-story steel moment-resisting frame.}
    \label{fig:37story}
\end{figure}

Within this setting, the wind-excited $37$-story steel moment-resisting framed building depicted in Fig.~\ref{fig:37story} is studied. The structure's height is 150 meters, with the ground floor having a height of 6 meters and all other floors measuring 4 meters in height. Each lateral load resisting frame consists of box section columns and wide flange beam sections selected from those suggested by the American Institute of Steel Construction. The structural steel of the members has a yield strength of 355 MPa and Young's modulus of 200 GPa. The structural mass includes the member self-weight and a carried mass calculated considering a non-structural building density of 100 kg/m$^3$. 

For this example, the response of one of the X-direction frames, as illustrated in Fig.~\ref{fig:37story}, is studied. The frame has six bays, each 6 meters wide, resulting in a total width of 30 meters. Rigid floor diaphragms are considered to act at each floor. As in the seismic example, OpenSees was used as the finite element modeling environment. In particular, displacement-based beam-column elements with a five-point Gauss–Legendre integration scheme were employed for modeling the structural members of the frame. Fiber discretization is performed for all sections, with elastic-perfectly-plastic constitutive law considered for all fibers. The Rayleigh damping matrix is calibrated in order to achieve an inherent damping of 2.5\% at the first two natural frequencies of the frame. To estimate the dynamic response of the frame, an adaptive average constant acceleration integration scheme with an initial time step of $0.02$ s was employed. The adaptive scheme tries various algorithms while decreasing the time step until convergence is achieved. 

%


\subsubsection{Stochastic Wind Load Model}
\label{S:Windmod}

The frame was subjected to wind excitation calibrated to a 10-minute mean wind speed of $v_H=65$ m/s at the building top with an associated wind direction of $\alpha=90^\circ$ (i.e., alongwind loading). Stochastic wind time series were generated through a data-driven spectral proper orthogonal decomposition (POD) scheme \cite{chen2005proper,chuang2019efficient,duarte2023uncertainty}. Wind tunnel data obtained from the Tokyo Polytechnic University (TPU) aerodynamic database \cite{TPU} was used for calibration purposes. The wind load representation was defined for the 3D building geometry, and the loads acting on the 2D frame in the X-direction were subsequently extracted. The wind tunnel data corresponded to a wind speed of 11 m/s at the top of a 1/300 rigid model of the building. Data was collected by 512 pressure taps with a sampling frequency of 1000 Hz. The collected pressure data was further processed to produce estimates of the forces at the center of each floor. Subsequently, spectral features were extracted by applying spectral POD to the floor load data. These features, which included the frequency-dependent eigenvalues and eigenvectors of the floor load vector, were then used to calibrate the data-driven spectral representation scheme outlined in \cite{chuang2019, duarte2023uncertainty}. One-sixth of the X-direction loads were considered to act on the frame under investigation and therefore define $\textbf{F}(v_H, \alpha)(t)$ for this application. To ensure reasonable initial and final conditions, the first and last minute of each wind load realization were linearly increased from zero and decreased to zero, respectively.


\subsubsection{Results}
\label{S:Windresults}

\begin{figure}[]
    \centering
    \includegraphics[width=\textwidth]{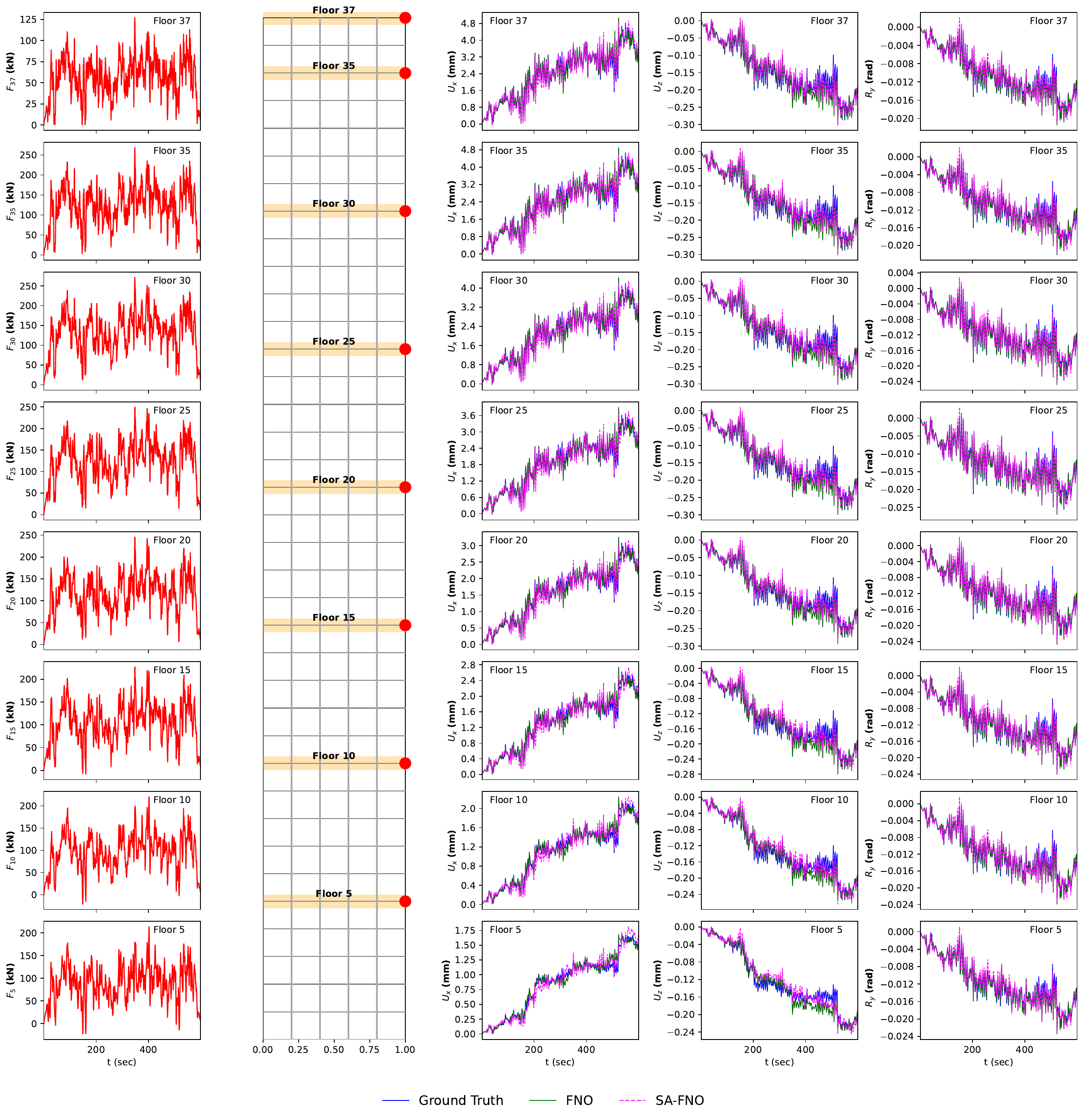}
    \caption{Comparison of the predictions made by the FNO and SA-FNO schemes. Results are shown only for the rightmost node (marked in red) for the highlighted floors.}   \label{fig:comparaison_fno_safno_deeponetfno}
\end{figure}

\begin{figure}[]
    \centering
    \includegraphics[width=\textwidth]{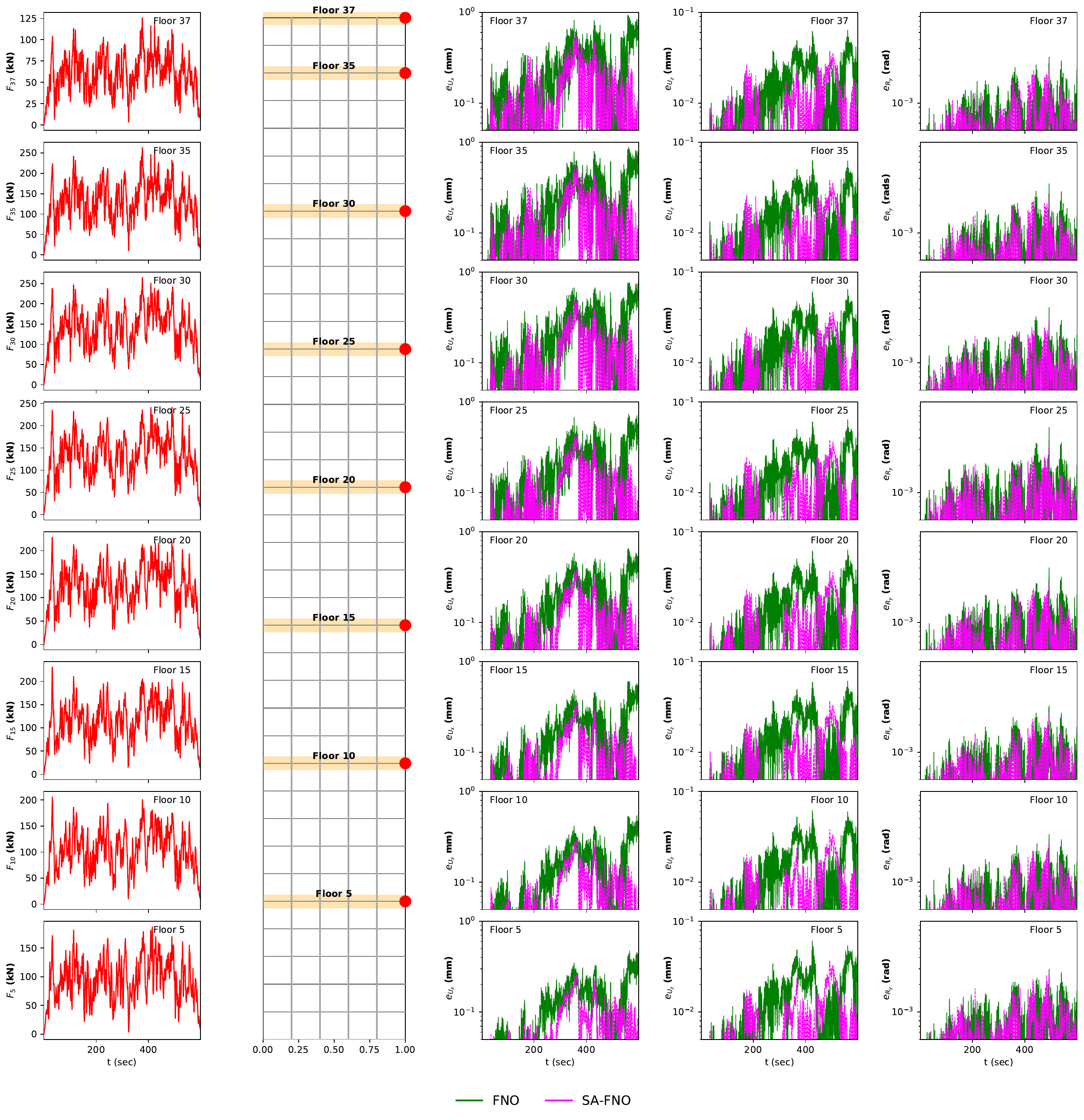}
    \caption{Accuracy comparison of the predictions of FNO and SA-FNO. Results are shown only for the rightmost node (marked in red) for the highlighted floors.}   \label{fig:comparaison_fno_safno_deeponetfno_error}
\end{figure}

Displacement responses at each degree of freedom were collected for each realization of the stochastic wind loads. A time step of $\Delta t = 0.02$ s was used in the response analysis, leading to a time history vector with 30000 time steps for each realization. To establish a robust mapping between the stochastic wind excitation and the structural response, 1000 model evaluations were performed with 800 evaluations dedicated to training and 200 reserved for testing. We removed the first 100 time points from both the input loading signal and the output response, as the loads were ramped from zero, resulting in negligible loading for the initial time steps. We, additionally, down-sampled the time history of the input forces and the output responses by a factor of 30. Fig.~\ref{fig:comparaison_fno_safno_deeponetfno} presents, for a representative sample, the responses in all three directions (two translations and one rotation) using FNO and SA-FNO for the rightmost node of the highlighted floors. Although both the FNO and SA-FNO show a good match, there are larger discrepancies in the prediction of FNO compared to SA-FNO, as shown by the error plots in Fig.~\ref{fig:comparaison_fno_safno_deeponetfno_error}. Leveraging the benefits of temporally dependent self-adaptive weights, SA-FNO can achieve higher accuracy than FNO for predictions at all floors of the X and Z translational responses, $U_{\text{X}}$ and $U_{\text{Z}}$, as well as the rotational response around the Y-axis, $R_{\text{Y}}$. 

\begin{figure}[]
    \centering
    \includegraphics[width=\textwidth]{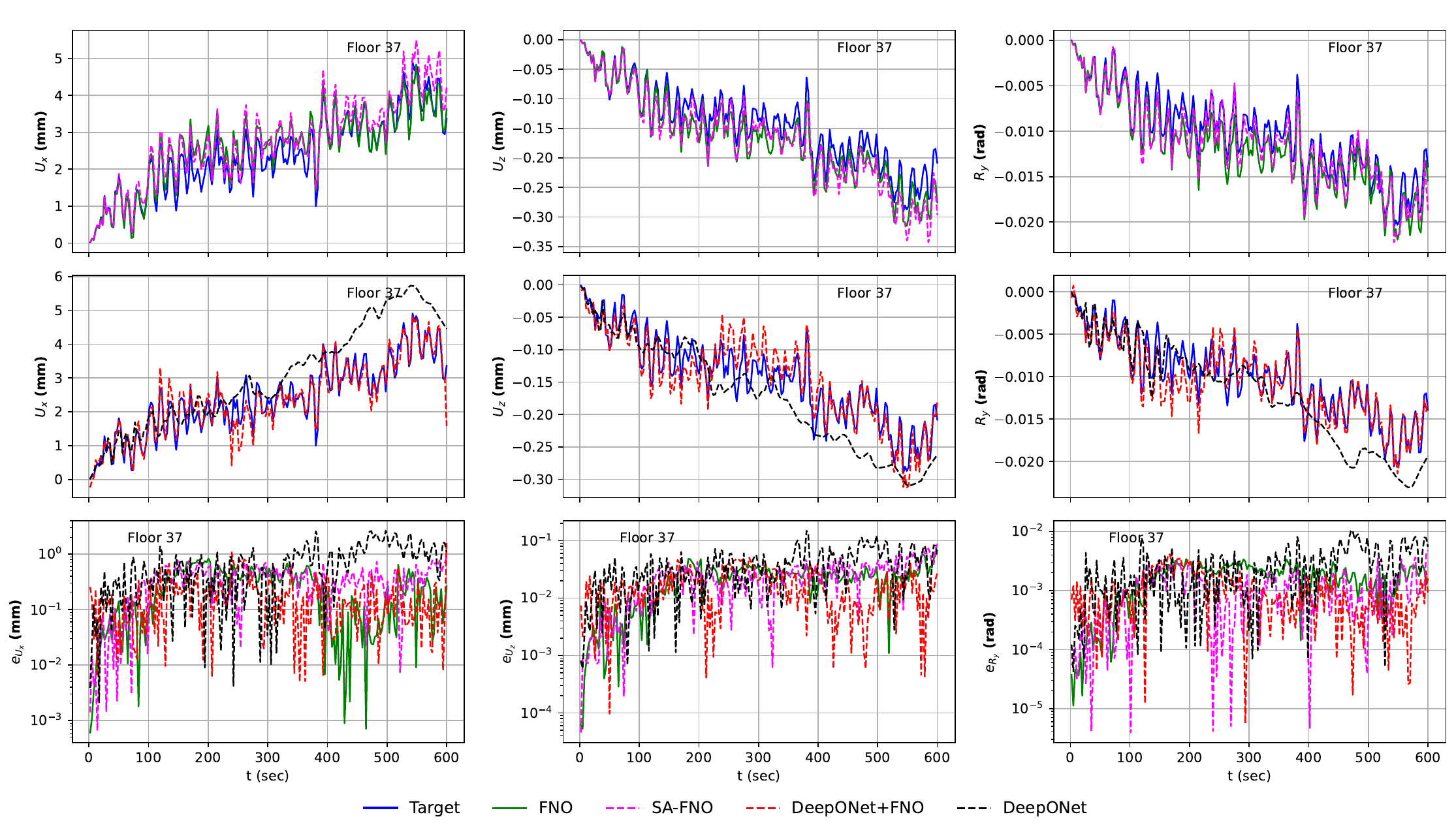}
    \caption{Comparative analysis of the performance of all tested neural operators in predicting the top floor wind response. The plot segregates the predictive capabilities of various NOs across different rows. The top row presents the FNO and SA-FNO, while the middle row displays DeepONet and DeepFNOnet. The bottom row illustrates the error profiles of all architectural variants, enabling a comparative assessment of their predictive accuracy.}
    \label{fig:comparaison_fno_safno_deeponetfno_sample}
\end{figure}

Again, we have employed the DeepONet only to model the responses to the wind loading at the top floor. Table~\ref{tab:performance_metrics_wind} presents the errors for each model evaluated for each directional output. We observe that FNO has the lowest errors for several error metrics. However, the plots of the representative sample shown in Fig.~\ref{fig:comparaison_fno_safno_deeponetfno_sample} demonstrate similar behavior of FNO and SA-FNO. Furthermore, we also observe that standalone DeepONet captures the overall trend of the response but cannot model the high-frequency content, while the response prediction is relatively accurate when using DeepFNOnet.

\begin{table}[]
\caption{Performance evaluation of all the NOs in predicting wind response of the top floor. Performance metrics are reported for all three response directions ($U_{\text{X}}$, $U_{\text{Z}}$, $R_{\text{Y}}$). Predictive accuracy of the NO architectures in terms of the MSE, best and worst sample performances, and relative $\mathcal{L}_2$ error metrics. The lowest error in each section is marked with \textbf{bold} text.}
\centering
\footnotesize
\begin{tabular}{llcccc}
\toprule
Direction & Model & MSE & Best Sample & Worst Sample & Relative \\
& & ($\times 10^{-1}$) & MSE ($\times 10^{-2}$) & MSE ($\times 10^{1}$) & $\mathcal{L}_2$ ($\times 10^{-1}$) \\
\midrule
\multirow{4}{*}{$U_{\text{X}}$} 
& FNO          & 5.763 & \textbf{2.898} & \textbf{10.96} & 1.887 \\
& SA-FNO       & \textbf{4.616} & 3.052 & 13.63 & \textbf{1.770} \\
& DeepFNOnet & 8.393 & 7.834 & 15.77 & 3.215 \\
& DeepONet     & 14.16 & 25.43 & 23.44 & 3.575 \\
\midrule
\multirow{4}{*}{$U_{\text{Z}}$}
& FNO          & \textbf{0.1925} & \textbf{0.9445} & \textbf{0.4292} & \textbf{1.899} \\
& SA-FNO       & 0.2253 & 1.590 & 0.4762 & 2.079 \\
& DeepFNOnet & 0.5091 & 8.600 & 0.8514 & 3.641 \\
& DeepONet     & 14.32 & 23.47 & 6.653 & 3.528 \\
\midrule
\multirow{4}{*}{$R_{\text{Y}}$}
& FNO          & \textbf{0.0008523} & \textbf{0.0004738} & \textbf{0.01801} & \textbf{1.790} \\
& SA-FNO       & 0.0009785 & 0.0007358 & 0.02318 & 1.963 \\
& DeepFNOnet & 0.001241 & 0.001127 & 0.02446 & 3.294 \\
& DeepONet     & 0.002606 & 0.006296 & 0.03699 & 3.769 \\
\bottomrule
\end{tabular}
\label{tab:performance_metrics_wind}
\end{table}


\section{Conclusions}
\label{S:conclusion}

This work introduces innovative extensions to neural operator frameworks for achieving high-fidelity predictions of the response trajectories of nonlinear structural systems subject to stochastic excitation modeling the effects of natural hazards. The proposed DeepFNOnet combines the strengths of DeepONet and FNO, yielding improved performance in capturing temporal dependencies and managing response prediction over long time horizons. The self-adaptive FNO further enhances predictive accuracy for systems subjected to high-frequency excitation. Case studies validate the robustness and generalizability of these architectures, achieving orders-of-magnitude faster computations than conventional models while maintaining high accuracy. Key outcomes include:
\begin{itemize}[leftmargin=*]
\item Computational Efficiency: Reduced simulation time by several orders of magnitude, enabling real-time forecasting of nonlinear response trajectories.
\item Predictive Accuracy: High precision in capturing seismic and wind-induced responses, with minimal deviation from ground-truth high-fidelity models.
\item Practical Applicability: Demonstrated applicability in scenarios ranging from multi-story seismic response analyses to dynamic wind response analyses of high-rise buildings.
\end{itemize}
These advancements pave the way for broader adoption of neural operators in civil engineering, significantly enhancing risk mitigation and infrastructure resilience for problems that require the rapid evaluation of response trajectories of systems subject to natural hazards whose time evolution is characterized through stochastic models.

\section*{Data Availability}
All codes will made available on GitHub, \url{https://github.com/Centrum-IntelliPhysics/Neural-Operators-for-Natural-Hazards}, upon publication of the paper.

\section*{Acknowledgments}
The authors' research efforts were partly supported by the National Science Foundation (NSF) under Grant No. CMMI-2118488 and No. TI-2140723 and Department of Energy (DOE) under Grant No. DE-SC0024162. The authors would like to acknowledge computing support provided by the Advanced Research Computing at Hopkins (ARCH) core facility at Johns Hopkins University and the Rockfish cluster. ARCH core facility (\url{rockfish.jhu.edu}) is supported by the National Science Foundation (NSF) grant number OAC1920103. Any opinions, findings, conclusions, or recommendations expressed in this material are those of the author(s) and do not necessarily reflect the views of the funding organizations.

\bibliography{Manuscript_R0}

\end{document}